\documentclass[lettersize,journal]{IEEEtran}
\usepackage{amsmath,amsfonts}
\usepackage{algorithmic}
\usepackage{algorithm}
\usepackage{array}
\usepackage[caption=false,font=normalsize,labelfont=sf,textfont=sf]{subfig}
\usepackage{textcomp}
\usepackage{stfloats}
\usepackage{url}
\usepackage{verbatim}
\usepackage{graphicx}
\usepackage{cite}
\usepackage{arydshln}
\usepackage{booktabs} 
\usepackage{pdfpages}
\usepackage{tabularx} 
\usepackage{multirow}
\usepackage{soul}
\usepackage{cancel}
\usepackage{xcolor}

\begin{document}

\title{KeyNode-Driven Geometry Coding for Real-World Scanned Human Dynamic Mesh Compression}

\author{Huong Hoang, Truong Nguyen, \textit{Fellow}, IEEE, and Pamela Cosman, \textit{Fellow}, IEEE 
\thanks{This work was supported by the Center for Wireless Communications, University of California, San Diego, and by the National Science Foundation under Grant DUE-1928604. }}

\maketitle

\begin{abstract}
The compression of real-world scanned 3D human dynamic meshes is an emerging research area, driven by applications such as telepresence, virtual reality, and 3D digital streaming. Unlike synthesized dynamic meshes with fixed topology, scanned dynamic meshes often not only have varying topology across frames but also scan defects such as holes and outliers, increasing the complexity of prediction and compression. Additionally, human meshes often combine rigid and non-rigid motions, making accurate prediction and encoding significantly more difficult compared to objects that exhibit purely rigid motion. To address these challenges, we propose a compression method designed for real-world scanned human dynamic meshes, leveraging embedded key nodes. The temporal motion of each vertex is formulated as a distance-weighted combination of transformations from neighboring key nodes, requiring the transmission of solely the key nodes' transformations. To enhance the quality of the KeyNode-driven prediction, we introduce an octree-based residual coding scheme and a Dual-direction prediction mode, which uses I-frames from both directions. Extensive experiments demonstrate that our method achieves significant improvements over the state-of-the-art, with an average bitrate savings of 58.43\% across the evaluated sequences, particularly excelling at low bitrates.
\end{abstract}


\section{Introduction}

The inherent depth and multitude of viewing angles enabled by 3D dynamic meshes contribute to an enriched experience across various applications, such as telepresence, virtual reality, and 3D digital streaming. Scanned 3D human dynamic meshes hold particular significance. They enable realistic representation and sharing of human motion, a capability increasingly accessible due to advancements in affordable scanning technologies. This accessibility is driving broader adoption, from entertainment and training simulations to healthcare and communication. However, the vast storage and transmission demands of these data-rich meshes necessitate efficient compression techniques designed to preserve the structural integrity and motion fidelity of scanned human meshes. 

Unlike synthesized dynamic meshes, such as those created using graphic design software, which typically feature fixed topology and well-defined structures, real-world scanned human meshes present unique challenges. Their topology often varies across frames, with changes in the number, connectivity, and correspondence of vertices. Furthermore, scanned meshes frequently suffer from defects such as holes, noise, and outliers, adding complexity to both prediction and compression. Developing methods that can effectively account for this temporal and spatial variability is a critical research challenge.

Human meshes also introduce challenges due to the combination of rigid motion (often for arms and legs) and non-rigid motion (often for the torso, fingers, and clothing). Unlike rigid objects, human motion involves intricate deformations and fine-grained movements that demand more sophisticated modeling to achieve accurate prediction and efficient compression.

This paper builds on our previous work \cite{10350716}, where we introduced a \textit{KeyNode-based} approach to describe the motion of evolving meshes using strategically placed key nodes. While our earlier research demonstrated the potential of using sparse key nodes to capture temporal motion in real-world scanned dynamic human meshes, this paper significantly extends the framework with new techniques that enhance both compression efficiency and temporal prediction. Specifically, we achieve better compression rates by explicitly quantizing KeyNode-driven motion vectors and optimizing entropy coding through the transfer of Huffman dictionaries fitted to Cauchy distributions. Moreover, we propose Octree-based residual coding to enhance geometry prediction accuracy and introduce a Dual-Direction prediction mode, enabling more adaptable modeling of dynamic human motion over time.  

Taken together with our prior contributions, these new advancements result in a comprehensive framework that effectively addresses the key challenges of compressing scanned 3D human dynamic meshes. Our codec offers the following key contributions:  

\begin{itemize}

    \item \textbf{A Novel Approach to Vertex Motion Modeling:} We propose a KeyNode-driven framework for modeling vertex motion, where each vertex is influenced by a weighted combination of transformations from a sparse set of key nodes. Unlike block-based methods, which assign a single rigid motion vector to entire spatial regions, our approach allows fine-grained, non-rigid deformation with per-vertex flexibility. This enables more accurate modeling of complex motion while maintaining a compact representation.

    \item \textbf{Tackling Topology Variability:} Instead of calculating the motion vector of individual vertices, which can be highly sensitive to changes in topology and scan defects as well as costly in bitrate, our method uses the motion of sparse controlling keynodes to represent vertex movement. This approach effectively captures the dynamics of the mesh, even with varying topology.
    \item \textbf{Efficient Handling of Human Motion:} To compute the KeyNode-driven motion vectors, the Keynode-Driven Codec uses a hybrid registration model that combines embedded and isometric deformation, capturing effectively both the rigid and non-rigid complexities of human motion, ensuring fidelity for intricate deformations in both anatomy and clothing.

    \item \textbf{Compression-aware Key Node Optimization:} Our work is the first to introduce a framework for selecting both the number and position of embedded key nodes in order to balance geometric fidelity with transmission cost. Prior embedded deformation work focused only on deformation quality, with no consideration of bitrate.
    \item \textbf{Residual-aware Octree Design:} Our KeyNode-driven prediction framework produces residuals that naturally cluster in localized surface regions. To exploit this structure, we propose an optimization scheme for constructing an unbalanced octree tailored to the spatial distribution of geometric residuals. Our approach dynamically balances bitrate and reconstruction accuracy by selectively expanding or pruning tree nodes. 

\end{itemize}

Overall, our codec effectively addresses the challenges inherent in compressing real-world scanned 3D human dynamic meshes, including topology variability and complex human motion. The experimental results show that our method significantly outperforms the state-of-the-art codec, particularly at low bitrates.

The remainder of this paper is organized as follows. Section II reviews related work on scanned dynamic mesh compression. Section III introduces our proposed codec, detailing the KeyNode-Driven compression framework, Octree-based residual coding, and Dual-direction prediction. Section IV presents the experimental setup and results. Finally, Section V concludes the paper and outlines future work.

\section{Related Work}

In this section, we review dynamic mesh compression methods.
Methods designed for meshes with fixed topology, where vertex correspondences are available across frames, rely on the topology remaining constant, allowing for predictable encoding of vertex motion \cite{10.5555/789086.789628, 10.1006/cviu.2002.0987,SPC, BICI2011577, 6387603, 10.1111:1467-8659.00433, Karni2004, 8486541, 10.1145/1073368.1073398, 10.1145/1028523.1028547, Payan2007}. For real-world scanned dynamic meshes, where the topology can change, and vertex correspondences are not inherently available, compression methods must accurately capture the temporal variations without explicit vertex correspondence. Previous methods can be organized into distinct categories.

\emph{Re-meshing-based methods:} Mesh frames with inconsistent topology can be mapped onto a uniform topology by re-meshing. In \cite{1464745}, an approximate global topology for the whole mesh sequence was constructed by re-meshing the first frame and mapping it to the following frames using motion estimation. Similarly, \cite{10.1145/2766945} employs separate Groups of Frames (GoFs), with each GoF having its own global topology.

\emph{Block matching-based methods:} Extending 2D video block matching, the method proposed in \cite{4358669} estimates temporal motion by block-wise comparisons. The 3D model is divided into block-based surfaces using cubic blocks, and their motions are estimated by searching for the best match block in the reference mesh frame. Similarly, a patch-based matching algorithm \cite{5652911} involves dividing frames into patches of the same surface area. For a patch in the current frame, the reference patch is determined as the one in the reference frame with the minimum dissimilarity.

\emph{Video-based methods:} The Visual Volumetric Video-based Coding (V3C) standard is used in \cite{9506298} to encode meshes with orthogonal projections, followed by atlas packing and video coding. Numerous proposals employing video-based methods were put forward in response to MPEG's CfP for Dynamic Mesh Compression \cite{MPEGCfP}. MPEG's benchmark codec, known as Anchor, involves initial geometry encoding of each frame with mesh decimation and Draco compression \cite{draco}. Texture frames are stacked into a video and encoded with standard video coding techniques.
 A patch-based approach was presented in \cite{9922890} reorganizing the intra-frame and inter-frame texture tiles to improve spatial and temporal correlation. In addition to mesh decimation, Mammou et al. proposed incorporating mesh subdivision, producing a base mesh and a series of subdivided meshes linked to their respective displacement vectors, and the temporal motion is captured by tracking vertex-level positional differences between consecutive base meshes \cite{9922888}. The displacements undergo wavelet transform, followed by quantization and packing into a 2D image or video, which is encoded by an image/video compression technique. MPEG adopted this approach \cite{9922888} for further development into a new standard, named Video-based Dynamic Mesh Coding (V-DMC).
Following its establishment, various efforts have been made to enhance V-DMC \cite{10222117, 10447762, 10402693, 10566382, 10772848, 10772781, 10647339, 10647545, 10647600, 10648035}.

\section{Proposed Codec}

As outlined in the Related Work section, many existing methods model temporal motion in dynamic mesh sequences by calculating per-block positional differences between successive frames. This strategy is conceptually similar to motion estimation in 2D video compression standards in H.264 or MPEG, where motion is often modeled using block-based techniques, assuming uniform, unidirectional motion across a region, with a single motion vector shared by all pixels within a block. Other existing methods get around this limitation by calculating per-vertex positional differences. However, this method would significantly increase the bit rate for transmitting vertex-level motion vectors.

Our approach, by contrast, adopts \textit{Embedded Deformation} (ED) to model \textit{vertex-level} motion in 3D space. ED enables smooth, non-rigid, and multi-directional motion by expressing each vertex’s movement as a weighted combination of transformations from a sparse set of key nodes. Because a small number of key nodes can efficiently encode complex deformations, this sparse key node approach can avoid the rate penalty of per-vertex motion information, but it provides more flexibility than the per-block methods, as each vertex can move independently without being bound to rigid spatial groupings, as in the case for block-based models.
Importantly, this approach tackles core challenges specific to scanned dynamic 3D meshes, particularly the absence of reliable vertex-to-vertex correspondence between frames in real-world 3D scans. Unlike 2D pixels, which maintain fixed grid positions and indices, vertices in dynamic meshes often vary in topology and ordering over time. This makes simple frame-to-frame differencing unreliable. Our deformation-based model overcomes these limitations, offering a more robust and generalizable solution for capturing temporal motion in dynamic 3D geometry.

Following our preliminary work in \cite{10350716}, which introduced the use of ED to model vertex-level motion in dynamic meshes, a recent paper \cite{10447762} also proposed using ED for motion modeling. While both works adopt ED as a core concept, their approach differs from ours in two important ways.
First, their usage of ED is limited to predicting the motion of a simplified base mesh generated from decimation, which is later subdivided to recover geometric detail. In contrast, we leverage the full flexibility of ED to model motion directly at the vertex level, capturing fine-grained deformations without relying on post-subdivision.
Second, as noted in their paper, their method is intended for use as a preprocessing step outside the core V-DMC codec. As such, it does not incorporate any bitrate constraints or optimize for compression. In contrast, our approach is explicitly designed for integration into the compression pipeline, so the design of the embedded deformation itself directly balances the competing goals of motion fidelity and transmission cost.

\begin{figure}[htb]
\centering
\includegraphics[width=0.485\textwidth]{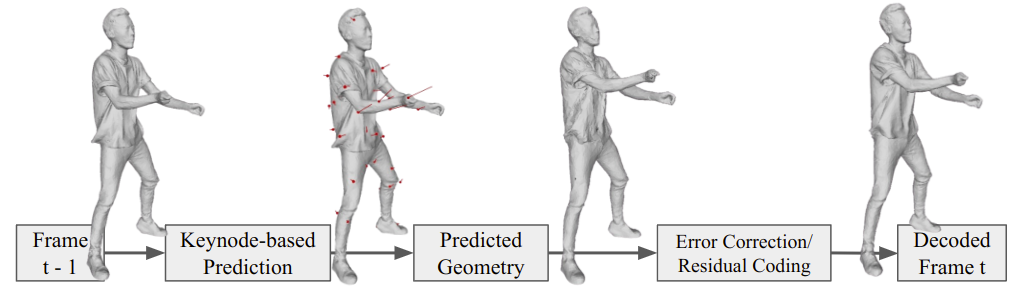}
\caption{Overall flow of the KeyNode-Driven approach.}
\label{fig:overall-structure}
\end{figure}

The process is depicted in Fig.~\ref{fig:overall-structure}. 
Starting with the geometry of frame $t-1$, we use the embedded key nodes to predict the geometry of frame $t$ through embedded deformation, where the geometry of frame $t-1$ is deformed to estimate the geometry of frame $t$. Since keynodes are typically sparse and the motion vectors are quantized, prediction errors arise.  To address these, we employ Octree-based residual coding to refine the predicted geometry for frame $t$. At the sequence level, we enhance the prediction process with a Dual-direction prediction mode. The details of the KeyNode-Driven coding are presented in Section \ref{sec:geo_coding}, residual coding in Section \ref{sec:res_coding}, and the Dual-direction prediction in Section \ref{sec:dual-direction}.

\subsection{KeyNode-Driven Geometry Coding}\label{sec:geo_coding}

Consider a dynamic mesh $S = \{M_1, M_2,..., M_F \}$, where $F$ represents the total number of frames in the sequence, and $M_t$ denotes a static mesh at time $t$ containing $V_{t}$ vertices. The average number of vertices is denoted $V$. Static meshes in $S$ may exhibit varying topologies.
Given two mesh frames $M_{t-1}$ and $M_{t}$, an approximation of $M_{t}$, denoted $M'_{t}$, can be obtained through deforming $M_{t-1}$ using an appropriate set of key nodes and their transformations. Let $\mathcal{GT}(.)$ be a geometric transformation operator which deforms $M_{t-1}$ to generate $M'_t$:

\begin{equation}
\label{eqn:mesh_transformation}
    M'_t = \mathcal{GT}(M_{t-1},\textbf{n}, \textbf{R}, \textbf{T})
\end{equation}
Here, $\textbf{n} =\{n_j \in \mathbb{R}^3 \} \in \mathbb{R}^{N \times 3} $ is the set of key nodes, $\textbf{R}  = \{R_j \in \mathbb{R}^3 \} \in \mathbb{R}^{N \times 3}$ and $ \textbf{T} = \{t_j \in \mathbb{R}^3 \} \in \mathbb{R}^{N \times 3}$ are the key nodes' transformations (rotations and translations), with $j \in [1, N]$ where $N<\!\!<V $ is the total number of key nodes controlling the deformation of the mesh.

\begin{figure}[htb]
\centering
\includegraphics[width=0.465\textwidth]{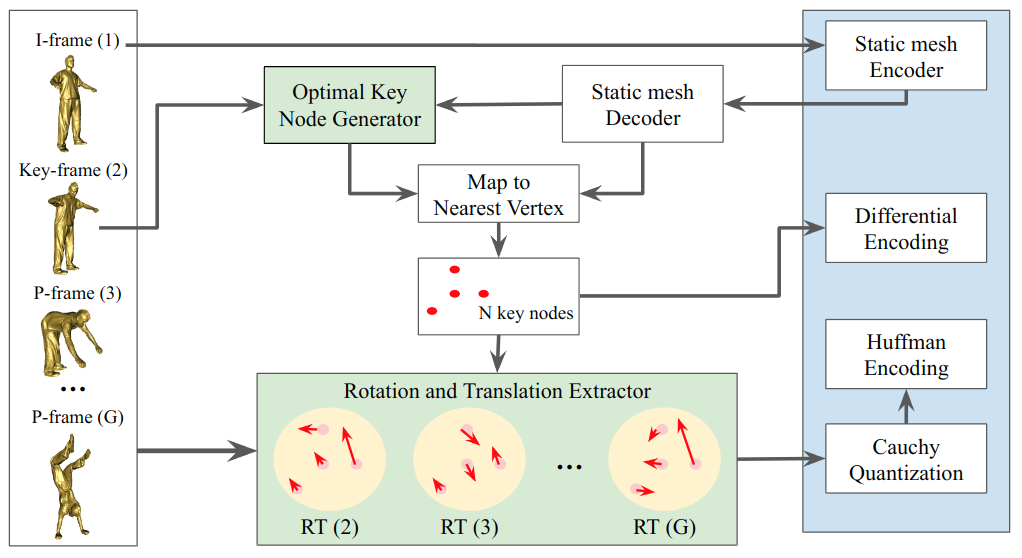}
\caption{Geometry encoder structure for a GoF with \textbf{\textit{G}} frames.}
\label{fig:encoder}
\end{figure}

Each vertex $x_i$ in $M_{t-1}$, $i \in [0, V_{t-1}-1]$  is deformed according to its $Q$ neighboring key nodes as:

\begin{equation}
\label{eqn:vertex_transformation}
    x_i' = \mathcal{GT}(x_i) = \sum_{j=0}^{Q-1} w_{ij} (R_j (x_i - n_j) + t_j + n_j)
\end{equation}
where $n_j, R_j, t_j$ are the position, rotation and translation of key node $j$, respectively, and $w_{ij}$ is the influence weight of key node $n_j$ for vertex $x_i$ based on the Euclidean distance between them.
To effectively compress, we need to reduce the number of nodes in $\textbf{n}$ and also identify $\textbf{R}$ and $\textbf{T}$ that capture the deformation between $M_{t-1}$ and $M_{t}$ using these nodes. Fig.~\ref{fig:encoder} illustrates the encoder structure. The \textit{Optimal Key Node Generator}, discussed below, aims to find a sparse set of key nodes $\textbf{n}$. The \textit{Rotation and Translation Extractor} aims to find the optimal transformations $\textbf{R}$ and $\textbf{T}$ for those nodes.

Our geometry encoder divides a dynamic mesh into GoFs, each containing an initial I-frame followed by several P-frames. I-frames are encoded using V-DMC \cite{9922888} all-Intra mode as a static mesh encoder as it consistently provided better performance than using Anchor for I-frame encoding. 

To determine an appropriate set of key nodes, we consider the decoded I-frame to be the \textit{source} and the subsequent P-frame (called the key-frame) to be the \textit{target} in a source-target deformation pair. All P-frames, including key-frames, are transmitted using transformations of key nodes.

The placement of key nodes plays a critical role in the quality of deformation. While ED has been widely adopted in geometry processing and animation tasks, previous works typically assume a fixed, predefined set of key nodes and focus solely on improving deformation accuracy. The key nodes are determined by uniform sampling over the mesh surface in \cite{10.1145/1276377.1276478} and \cite{Chen2022}, while \cite{10447762} uses the base mesh, a decimated version of the original mesh. In the context of compression, however, it is essential to also consider \textit{bitrate}, introducing a new trade-off between geometric fidelity and representation cost.

To tackle this, our earlier work \cite{10350716} introduced a \textit{Key Node Generator}, which formulates an optimization framework to jointly balance deformation quality and transmission efficiency. A central challenge here is that incorporating bitrate, which ties closely to the number of key nodes, directly into the standard ED optimization is problematic, as ED traditionally assumes the key nodes are fixed when solving for the local transformations ($\textbf{R}$ and $\textbf{T}$). Our method circumvents this by alternating between estimating deformations, pruning key nodes, and adjusting key node positions. This allows the layout of key nodes to adapt dynamically based on both the underlying motion and the associated encoding cost.
To the best of our knowledge, this is the first approach to integrate ED with a compression-aware optimization of the key node configuration, enabling a more efficient representation of dynamic 3D geometry. 
The \textit{Optimal Key Node Generator} block in Fig.~\ref{fig:encoder} employs that method, ensuring the set of chosen key nodes meets the target sparsity while minimizing distortion.

Once the optimal key nodes are identified, we map them to the decoded I-frame to find the closest vertex for each key node. These closest vertex positions are then adopted as the finalized positions of the key nodes, and will be used as $n_j$ in Eq.~\ref{eqn:vertex_transformation}. We sort these vertex indexes in ascending order and encode them using differential encoding.

The optimal values for $\textbf{R}$ and $\textbf{T}$ are obtained through an optimization process in the \textit{Rotation and Translation (RT) Extractor} module in Fig.~\ref{fig:encoder}, formulated as:

\begin{equation}
\label{eq:ori_deformation_operator}
 \underset{\textbf{R, T}}{\mathrm{argmin}} (\mathcal{L}_{data} + \alpha_{reg} \mathcal{L}_{reg} )
\end{equation}
\noindent Here, $\mathcal{L}_{data}$ penalizes the geometry deviation between source and target, and $\mathcal{L}_{reg}$ ensures the smoothness of the deformation. We employ the approach introduced in \cite{Chen2022} to solve this optimization problem.
Note that the approach in \cite{Chen2022} is designed to maximize deformation quality without considering compression constraints. To this end, it employs a densely sampled set of key nodes distributed uniformly over the mesh surface. In contrast, our work addresses the fundamentally different challenge of compression, where bitrate plays a major role. Rather than using a dense and fixed key node layout, we replace it with a sparse, optimized key node set generated by our previously introduced \textit{Optimal Key Node Generator}. This compression-aware design balances accuracy with representation cost, enabling efficient and scalable mesh prediction.

\begin{figure}[htb]
\centering
\includegraphics[width=0.49\textwidth]{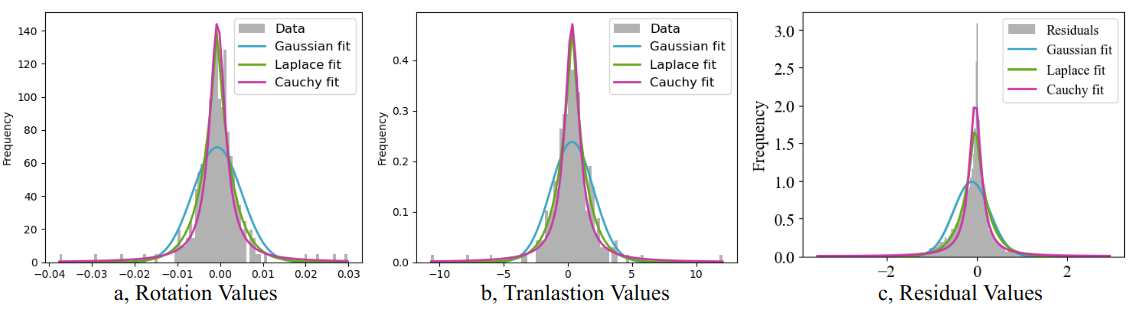}
\caption{Distributions of RT and Residuals for frame 619 of \textit{Thomas}. The histograms are plotted with 200 bins.}
\label{fig:distributions}
\end{figure}

For each P-frame, keynode transformations including $\textbf{R} \in \mathbb{R}^{N \times 3}$ and $\textbf{T} \in \mathbb{R}^{N \times 3}$ are flattened into a rotation vector and translation vector, denoted $\textbf{R}^f \in \mathbb{R}^{3N} $ and $\textbf{T}^f \in \mathbb{R}^{3N}$. Fig.~\ref{fig:distributions} presents the distributions of our data with the fitted Gaussian, Laplace, and Cauchy distributions. Among these, the Cauchy distribution demonstrates the closest alignment with the observed data, capturing both the sharp peak and heavy tails more accurately than the others. This insight plays a key role in our compression strategy. By modeling our transformations with a Cauchy distribution, we are able to design a quantization scheme that is statistically well-matched to the data. Crucially, this also allows for the implicit construction of the Huffman coding table at the decoder—meaning the coding dictionary does not need to be transmitted explicitly. This eliminates additional overhead and enhances compression efficiency.

Two Cauchy distributions are fitted to $\textbf{R}^f$ and $\textbf{T}^f$. With a specified codebook size $N^{RT}_{b}$, $N^{RT}_{b}$ equal-width bins are generated spanning from $-|b|$ to $|b|$, where $b$ is the maximum absolute value within a given frame. The two middle bins are merged to create a dead-zone.
The bin $(b_1^k, b_2^k]$ has its quantized value as the expected value of the Cauchy probability density function (pdf) within that range:

\begin{equation}
\label{eq:EX_cauchy}
   \Theta_k =  \int_{b_1^k}^{b_2^k} \frac{x}{\pi \gamma ( 1 + (\frac{x - x_0}{\gamma})^2)} \,dx
\end{equation}
where $x_0, \gamma$ are the mean and variance of the fitted Cauchy distribution. Each transformation value $l$ is quantized to the expected value of the range that it belongs to: $\hat{l} = \Theta_k  $, if $b_1^k < l < b_2^k$.

A Huffman table is generated based on the probabilities; the probability of bin $k$ with range $(b_1^k, b_2^k]$ is:

\begin{equation}
\label{eq:pk_cauchy}
   P_k =  \int_{b_1^k}^{b_2^k} \frac{1}{\pi \gamma ( 1 + (\frac{x - x_0}{\gamma})^2)} \,dx
\end{equation}

Hence, the overall transmission cost for each $\textbf{R}^f$ or $\textbf{T}^f$ contains the following components: mean and variance of the fitted Cauchy distribution, the border value $|b|$ for quantization bins, and the Huffman encoded bitstream.

\begin{figure}[htb]
\centering
\includegraphics[width=0.465\textwidth]{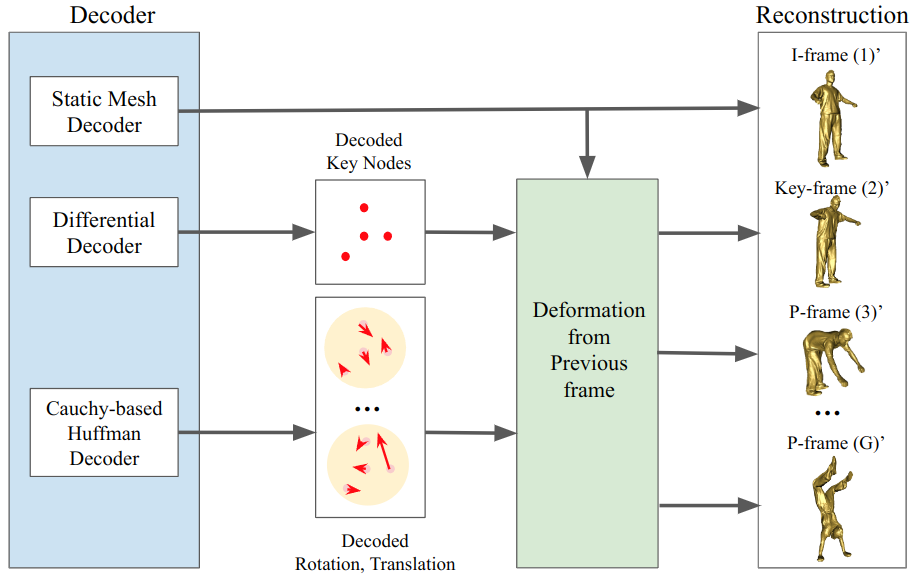}
\caption{Geometry Decoder structure for a GoF with \textbf{\textit{G}} frames}
\label{fig:decoder}
\end{figure}

At the decoder (Fig.~\ref{fig:decoder}), I-frames are decoded with the static mesh decoder.
Given $\textbf{n}, \textbf{R}, \textbf{T}$ retrieved through differential and Huffman decoding
for P-frame $M_t$, its distorted mesh $M'_t$ is reconstructed via deformation from $M'_{t-1}$, using Eq.~\ref{eqn:vertex_transformation}.

\subsection{Octree-based Geometric Residual Coding}\label{sec:res_coding}

Residual coding is a critical component of any predictive compression method. Unlike prior mesh compression approaches such as V-DMC—which flatten 3D vertex-level residuals into 2D images and encode them using conventional video compression techniques—our KeyNode-based prediction framework operates natively in 3D space. We observed that the geometric residuals naturally cluster in localized surface regions, due to the spatial smoothness and locality of the deformation model.

This spatial coherence of residuals motivates the proposed octree-structured residual coding. The octree structure allows residuals to be encoded at varying spatial resolutions, adapting to the complexity of different regions. Importantly, we do not use a standard balanced octree. Instead, we propose an optimization approach to determine an optimal \textit{unbalanced octree}, which considers both correction quality and bitrate during construction.

\subsubsection{Spatial Octree-structured Residual Quantization}
\label{subsection:leaf-res-quantization}

\begin{figure}[htb]
\centering
\includegraphics[width=0.47\textwidth]{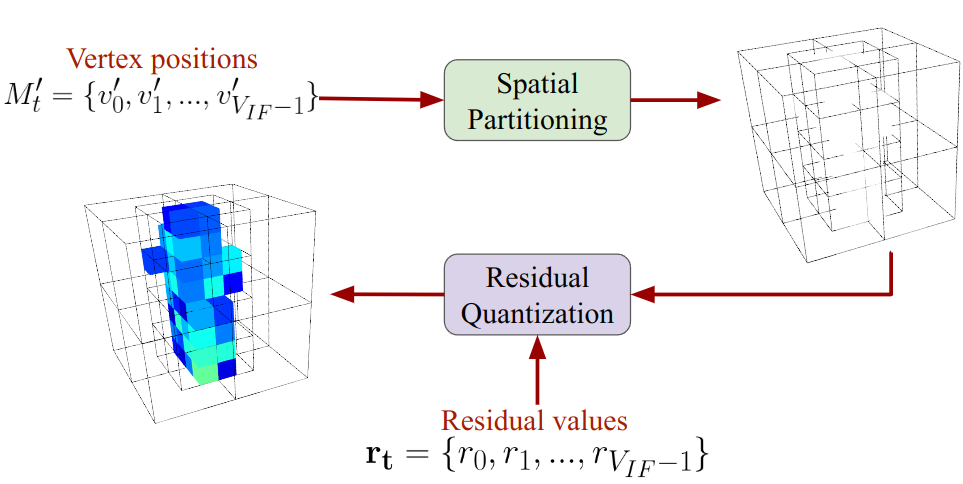}
\caption{Spatial octree-structured residual quantization method. Cell colors represent quantized residual values.}
\label{fig:residual-octree-quantization}
\end{figure}

\begin{figure}[htb]
\centering
\includegraphics[width=0.47\textwidth]{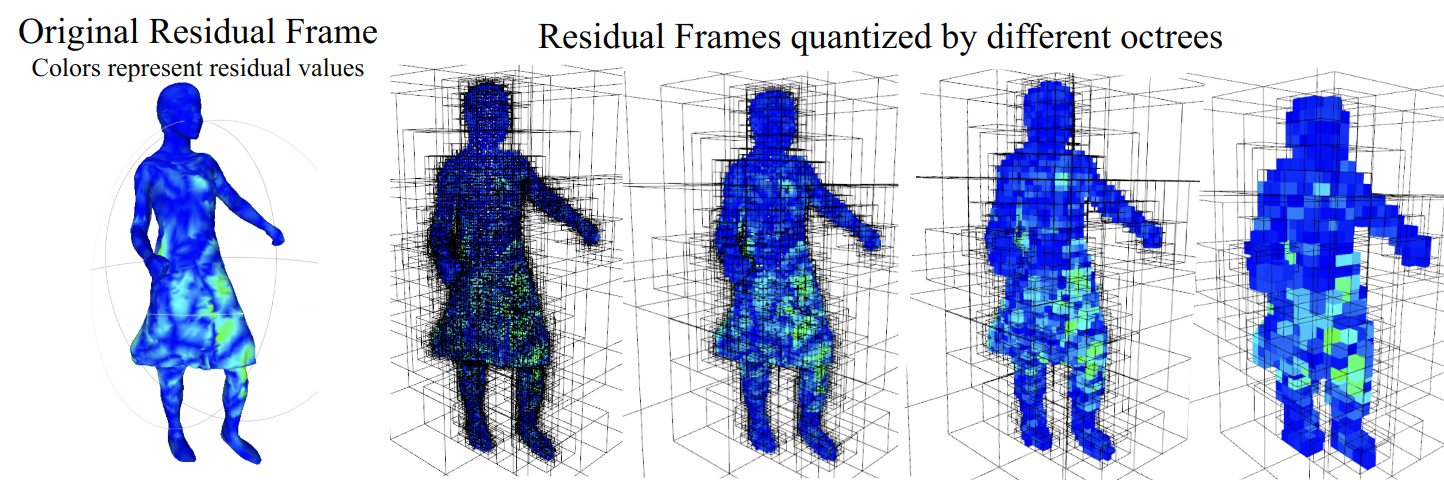}
\caption{Visualizations of a residual frame being quantized by different octrees, resulting in different levels of detail.}
\label{fig:octree-examples}
\end{figure}

Predicting P-frames through the transformation of embedded key nodes often means that areas of high motion or intricate detail exhibit larger residuals than those with slow and rigid motion.
Let $M'_t = \{v'_0, v'_1,..., v'_{V_{IF} -1} \}$ denote the reconstructed mesh at time $t$, where $v'_i$ represents the distorted position of vertex $i$, and $V_{IF}$ is the number of vertices in $M'_t$ (also in the I-frame associated with $M_t$).  Let $\mathbf{r_t} = \{r_0, r_1,..., r_{V_{IF} -1} \}$ be the residual frame which represents the displacement of $M'_t$ vertices relative to the source mesh $M_t$.
To take advantage of the residual clustering effect, the \textit{Spatial Partitioning} module (see Fig.~\ref{fig:residual-octree-quantization}) divides vertices in $M'_t$ into octree cells based on their positions. The resulting octree contains $L$ non-empty leaf nodes, $\mathcal{L} = \{l_0, l_1,..., l_{L-1}\}$, where leaf node $l$ contains $V_{l}$ vertices $\{v'^l_{0}, v'^l_{1},..., v'^l_{V_{l} -1}\} \subset M'_t$ that have residuals $\{r^l_{0}, r^l_{1},..., r^l_{V_{l} -1}\}$, a subset of $\mathbf{r_t}$. In the \textit{Residual Quantization} module, the quantized value $\hat{r}^l$ for these residuals is computed as the average of the residuals belonging to non-empty leaf node $l$. 
Fig.~\ref{fig:octree-examples} shows some residual frames quantized by different octrees.
The system transmits the octree structure and $\hat{r}^l$ for each leaf. 

As P-frames are deformed from their I-frame and inherit the I-frame's topology, calculating prediction residuals is challenging since the reconstructed P-frames lack vertex correspondence with the source P-frames. A common technique to calculate $\mathbf{r_t}$ is to compute the position difference between each distorted vertex and its closest vertex in the source mesh. However, if multiple distorted vertices share the same closest source vertex, perfect residual correction would lead them to all move to the same position in the corrected mesh. This issue, called duplicated vertices, can lead to undesired self-intersection of triangles in the corrected mesh. To avoid this, instead of mapping $v'_i$ to the closest vertex in $M_t$, we identify the closest point on the surface of $M_t$, denoted $v^*_i$. The residual of a vertex $i$ is then calculated as the positional difference between $v^*_i$ and $v'_i$:
\begin{equation}
    r_i = v^*_i - v'_i
\end{equation}

The mesh containing vertices \{${v_i^*}$\} and having $M'_t$'s topology is called the \textit{approximate source mesh} for $M'_t$.

\subsubsection{Cost-constrained Octree Generation}
\label{subsection:octree-gen}
Determining a suitable octree structure to efficiently quantize the residuals is crucial. In this section, we present a cost-constrained optimization approach for generating a suitable unbalanced octree for each residual frame, balancing residual distortion and octree size (bitrate) through a cost function of an optimization problem.
When a residual frame $\mathbf{r_t}$ is quantized by an octree $T$, we denote the quantized frame by $\mathbf{\hat{r}_t} = \{\hat{r}_0, \hat{r}_1,..., \hat{r}_{V_{IF} -1} \}$, where each $\hat{r}_i$ represents the average residual of the leaf node to which $r_i$ belongs. The overall quantization distortion can be calculated as the mean squared error between the original and quantized residual frames:
\begin{equation}
    D(T) = MSE (\mathbf{r_t}, \mathbf{\hat{r}_t}).
\end{equation}

Let $c(T)$ be the cost of octree $T$; we link $c(T)$ to the number of leaf nodes, since the size of $T$ directly affects the bitrate.
Given a cost constraint $c$, the octree with minimized distortion is called the optimal tree for constraint $c$:

\begin{equation}
    T_{opt} = argmin D(T)
\end{equation}
such that $c(T) \leq c$. To identify $T_{opt}$, we compute a `quality factor' $\lambda$ for each node $n$ based on the ratio of the change in distortion to the change in cost from expanding node $n$:
\begin{equation}
\label{eq:node_lambda}
    \lambda(n) = \frac{\Delta D(n)}{\Delta C(n)} = \frac{D(T) - D(T(n))}{C(T(n)) - C(T)}
\end{equation}
Here, $D(T)$ is the distortion of the octree $T$ in which $n$ is a leaf node, $D(T(n))$ is the distortion of the octree in which $n$ is expanded, and
$\lambda(n)$ is the slope of the cost-distortion curve when adjusting the octree structure. 
When growing the octree, the node with the largest  $\lambda(n)$  will be chosen to expand, and when pruning, the node with the smallest $\lambda(n)$ will be made a leaf. 
To avoid calculating the distortion and cost for the whole tree, we compute $\lambda(n)$ based on the distortion of the node being evaluated \cite{Lin1994}:
 \begin{equation}
     \Delta D(n) = D(n) - \sum_{\i=0}^{7} \frac{p(n_i)}{p(n)} D(n_i)
 \end{equation}
Here, $n_i, 0 \leq i \leq 7$, are the possible children of node $n$, $p(n) = \frac{V_{n}}{V_{IF}}$, where $V_{n}$ is the number of vertices in node $n$ and $D(n)$ is the distortion between the original and quantized residuals in node $n$:
\begin{equation}
    D(n) = MSE \Bigl( \{r^n_{0}, r^n_{1},..., r^n_{V_{n} -1}\}, \{\hat{r}^n, \hat{r}^n,..., \hat{r}^n \} \Bigr)
\end{equation}
where $\hat{r}^n$ is the quantized residual value for node $n$.

\subsubsection{Coding of Octree-based Quantized Residuals}

\begin{figure}[htb]
\centering
\includegraphics[width=0.5\textwidth]{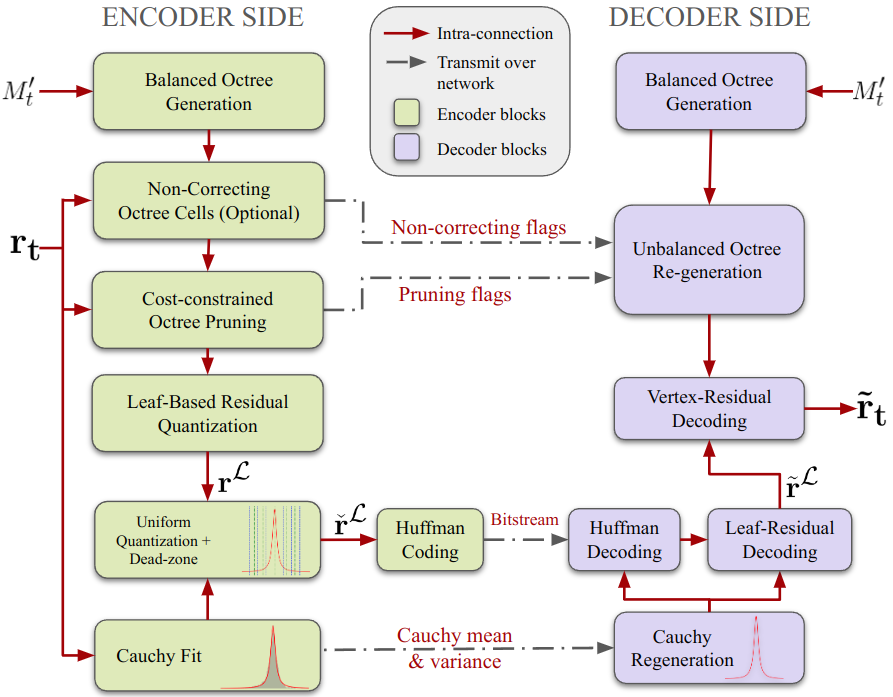}
\caption{Codec structure for Residual coding.}
\label{fig:residual-codec-structure}
\end{figure}

Fig.~\ref{fig:residual-codec-structure} depicts the structure for residual coding.
At the encoder, the distorted P-frame $M'_t$ is used to generate a balanced octree based on the vertex positions. 
When vertices are accurately predicted through deformation, they should not be corrected with residual coding. To determine which nodes should carry residuals, we compare the average vertex distortion $e^l$ of the node's vertices with a threshold $\mathcal{T}$. In our implementation, the value of $e^l$ is derived from the Mesh Structural Distortion Measure ver.~2 (MSDM2) \cite{Lavou2011} distortion metric, while it could also be directly calculated as $\hat{r}^l$ to save computational cost. All nodes where $e^l < \mathcal{T}$ are referred to as Non-Correcting Octree Cells (NCOC); residuals for their vertices will not be transmitted. An octree cell is also classified as an NCOC if its local MSE distortion after residual correction is greater than or equal to the MSE distortion before correction.
Non-correcting flags are sent only for the highest-level NCOC parents, as their children inherit the same type.
The octree is then pruned using the cost-constrained optimization method. Pruning flags are sent to the decoder, allowing it to reconstruct the unbalanced octree. We also disable residual coding for the entire frame if the distortion increases after correction.

The residuals are quantized with the unbalanced octree, and then they undergo Cauchy-fitted uniform quantization with a dead-zone, similar to the way the R and T transformations are handled. A Cauchy distribution is fitted to the original residual frame $\mathbf{r_t}$ in the corresponding dimension, which is used to construct the Huffman codebook in a manner similar to the process described in Eqs.~\ref{eq:EX_cauchy} and \ref{eq:pk_cauchy}. Overall, transmitting the residual frame $\mathbf{r_t}$ requires: two binary streams of flags, a bitstream of Huffman-coded leaf residuals, and a fitted Cauchy distribution described by its mean and variance.

To decode the residual frame $\mathbf{r_t}$, the unbalanced octree is constructed using the distorted P-frame $M'_t$ and the received flags. Leaf residuals $\Tilde{\textbf{r}}^\mathcal{L}$ are obtained by decoding bitstreams using the Huffman codebook constructed through the received  Cauchy distributions. The decoded vertex residual $\Tilde{r}_i \in \mathbf{\Tilde{{r}}_t}$ is calculated as the value of the quantized residual of its associated leaf.

Although our residual estimation process is similar to that of V-DMC—both compute residuals by projecting onto the original mesh surface—our method fundamentally diverges in how residuals are quantized and encoded. V-DMC applies a wavelet transform to the residuals (referred to as \textit{displacements}), quantizes the coefficients, and encodes them by traditional video compression techniques after packing them into a 2D video.
In contrast, our method maintains the intrinsic 3D nature of the data throughout the compression pipeline. We perform quantization directly in 3D space, leveraging an adaptive octree structure that aligns with the spatial distribution and complexity of the residuals. For entropy coding, we exploit the nature of the residuals' distribution using a Cauchy-based coding scheme, enabling compact and statistically informed representation. By staying within the 3D domain, our method preserves geometric fidelity and achieves efficient compression.

\subsection{Dual-direction Prediction}\label{sec:dual-direction}

\begin{figure}[htb]
\centering
\includegraphics[width=0.45\textwidth]{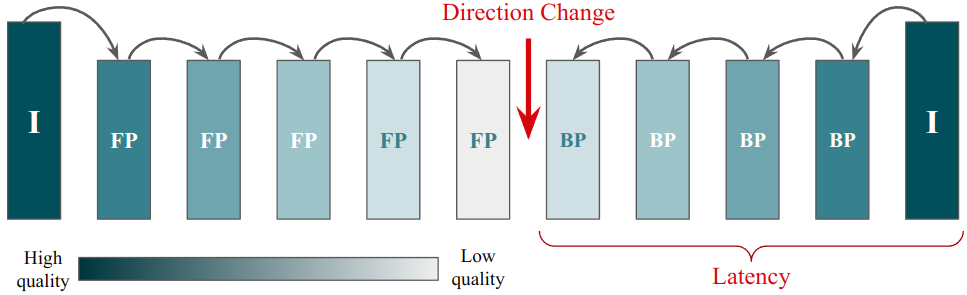}
\caption{Dual-direction Prediction flow, where P-frames are predicted from both directions. Labels: \textbf{I} for I-frames, \textbf{FP} for Forward P-frames, and \textbf{BP} for Backward P-frames.}
\label{fig:dual-direction-prediction}
\end{figure}

In our codec, P-frames are predicted from either an I-frame or another P-frame. Distortion can accumulate across frames, and later P-frames typically have higher distortion than earlier ones.
Dual-direction Prediction (Fig.~\ref{fig:dual-direction-prediction}) predicts P-frames from both directions. A \textit{switch index} determines when the prediction direction changes. 

If the switch index is predetermined and fixed, the prediction direction is based solely on temporal distance from the reference frame, ensuring minimal complexity. This results in a fixed number of backward P-frames, and the latency is equal to this number. 

When the switch index is determined adaptively, the codec evaluates prediction quality in both forward and backward directions for each P-frame to identify the optimal switching point.  In this mode, temporal distance alone does not dictate the prediction direction; instead, quality considerations drive the choice. Adaptive switching increases latency to the size of the GoF and requires additional encoder storage to hold frames predicted from both directions until the switch index is finalized.

The selection of fixed or adaptive direction switching depends on the application, especially concerning latency and encoder storage capacity.

\section{Evaluation}

\subsection{Datasets and Metrics}

To thoroughly evaluate the performance of our KeyNode-driven codec, we conducted experiments on a diverse set of scanned dynamic mesh sequences, each chosen to highlight different challenges and characteristics. Details about the mesh sequences are in Table \ref{tab:dataset-details}.
We categorize the sequences into three groups. In the \textit{Slow Motion} category, consecutive frames usually have significant overlap. In the \textit{Fast Motion} category, consecutive frames often exhibit noticeable changes in the position or orientation of body part(s) and/or rapid movement of clothing. \textit{Join/Split} contains a mesh sequence with dynamic interactions between two objects (a person and a ball) that connect and disconnect across frames.

\begin{table}[htb]
\caption{Statistics of the evaluated datasets.}
\resizebox{0.488\textwidth}{!}{
\begin{tabular}{|c|c|c|c|c|c|c|}
\hline
Category                                                               & Sequence   & \begin{tabular}[c]{@{}c@{}}First\\ Frame\end{tabular} & \begin{tabular}[c]{@{}c@{}}Last \\ Frame\end{tabular} & \begin{tabular}[c]{@{}c@{}}Avg. \#\\ Vertices\end{tabular} & \begin{tabular}[c]{@{}c@{}}Avg. \# \\ Faces\end{tabular} & \begin{tabular}[c]{@{}c@{}}Capture \\ Artifacts\end{tabular} \\ \hline
\multirow{3}{*}{\begin{tabular}[c]{@{}c@{}}Slow\\ Motion\end{tabular}} & Mitch      & 1                                                     & 300                                                   & 16676                                                      & 30000                                                    &                                                              \\ \cline{2-7} 
                                                                       & Soldier    & 536                                                   & 835                                                   & 22702                                                      & 39989                                                    & Holes                                                             \\ \cline{2-7} 
                                                                       & Thomas     & 618                                                   & 917                                                   & 16235                                                      & 30000                                                    &                                                              \\ \hline
\multirow{3}{*}{\begin{tabular}[c]{@{}c@{}}Fast\\ Motion\end{tabular}} & Longdress  & 1051                                                  & 1350                                                  & 22231                                                      & 39991                                                    & Outliers  \& holes                                                    \\ \cline{2-7} 
                                                                       & Dancer     & 1                                                     & 300                                                   & 20908                                                      & 39373                                                    &   Holes                                                           \\ \cline{2-7} 
                                                                       & Levi       & 0                                                     & 149                                                   & 47580                                                      & 39984                                                    & Holes                                                             \\ \hline
Join/Split                                                             & Basketball & 1                                                     & 300                                                   & 20903                                                      & 39456                                                    & Outliers \& holes                                                    \\ \hline
\end{tabular}}
\label{tab:dataset-details}
\end{table}

In this article, we use MSDM2 \cite{Lavou2011} for geometry distortion evaluation due to its proven high correlation with human visual perception through extensive subjective experiments. MSDM2 measures local curvature differences between two meshes, providing a distortion score in the range [0,1], where 0 indicates identical meshes, and values approaching 1 signify substantial visual dissimilarity. Since MSDM2 is designed for static meshes, to determine the overall sequence distortion, the average value across frames is calculated.

\subsection{KeyNode-based Compression Hyper-parameters}

The number $N^{res}_{b}$ of quantization levels for residuals is varied across  $2^7, 2^8$, and $2^9$.  The codebook size $N^{RT}_{b}$ for key node rotations and translations is varied across $2^3, 2^5$, and $2^7$. The size of the unbalanced octree for Residual coding varies across 20, 100, and 500 leaf nodes. The GoF size varies across 2, 4, 8, 16, and 32. We assigned 100 key nodes to \textit{Soldier}, \textit{Thomas}, \textit{Mitch}, and \textit{Levi}, and 300 to \textit{Dancer}, \textit{Longdress}, \textit{Basketball}. More key nodes are assigned to sequences with more complex motion to enhance P-frame prediction.

We evaluate two variants of our KeyNode-based method: \textit{Fully Forward (FF)}, where each P-frame is predicted from its preceding frame, and \textit{Adaptive Dual-direction Prediction (ADP)}, where the change in direction depends on the prediction quality from both directions, leading to a latency equal to the GoF size.

\subsection{Analysis of KeyNode-based Encoded Components}

\begin{figure}[h]
\centering
\includegraphics[width=0.45\textwidth]{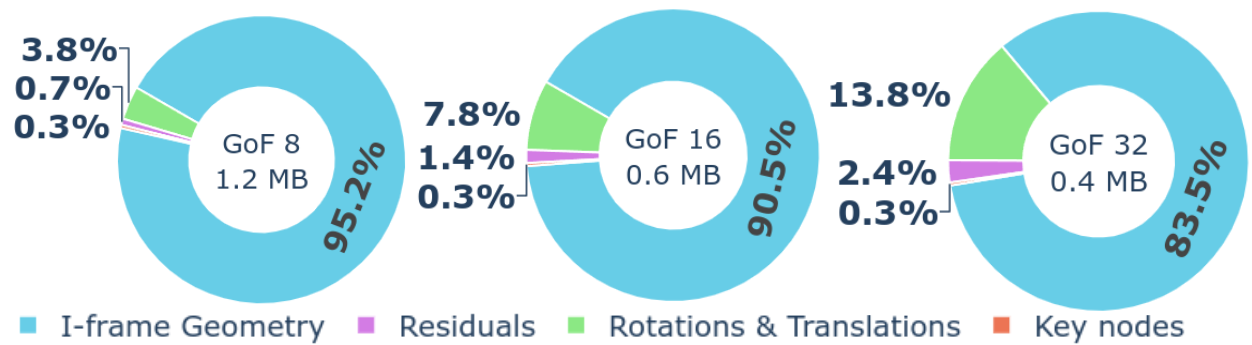}
\caption{Encoded components when varying GoF size in \textit{Thomas}. The number of key nodes is 100.}
\label{fig:pie-chart-100-nodes}
\end{figure}

\begin{figure}[h]
\centering
\includegraphics[width=0.45\textwidth]{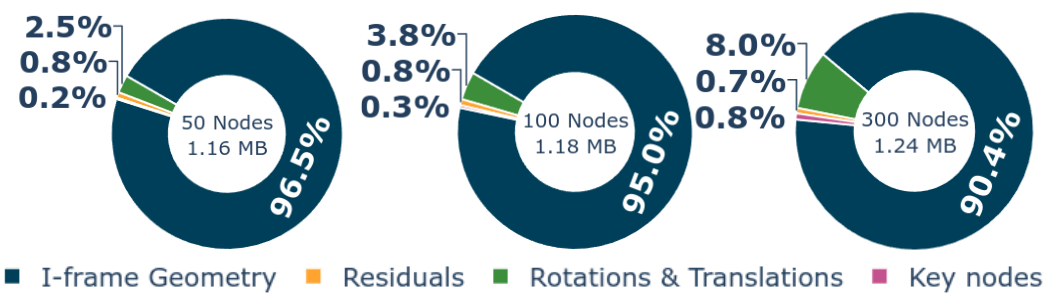}
\caption{Encoded components when varying number of key nodes in \textit{Thomas}. GoF size is 8 frames.}
\label{fig:pie-chart-GoF5}
\end{figure}

This section offers a closer examination of our codec elements. For a dynamic mesh, we transmit the statically encoded I-frames, encoded key nodes, encoded RT, and encoded residuals for P-frames. 
Figs.~\ref{fig:pie-chart-100-nodes} and \ref{fig:pie-chart-GoF5} illustrate the proportion of each encoded component as the GoF size and number of key nodes vary. The proportion of encoded residuals increases the most with GoF size increases; this is due to a smaller number of I-frames for large GoF cases. When the number of key nodes increases, the encoded RT component increases proportionally, while the geometry residual proportion slightly decreases, because a larger number of key nodes does better P-frame prediction, leading to lower prediction errors.

\subsection{Geometry Errors at each Compression Stage}

\begin{figure}[htb]
\centering
\includegraphics[width=0.45\textwidth]{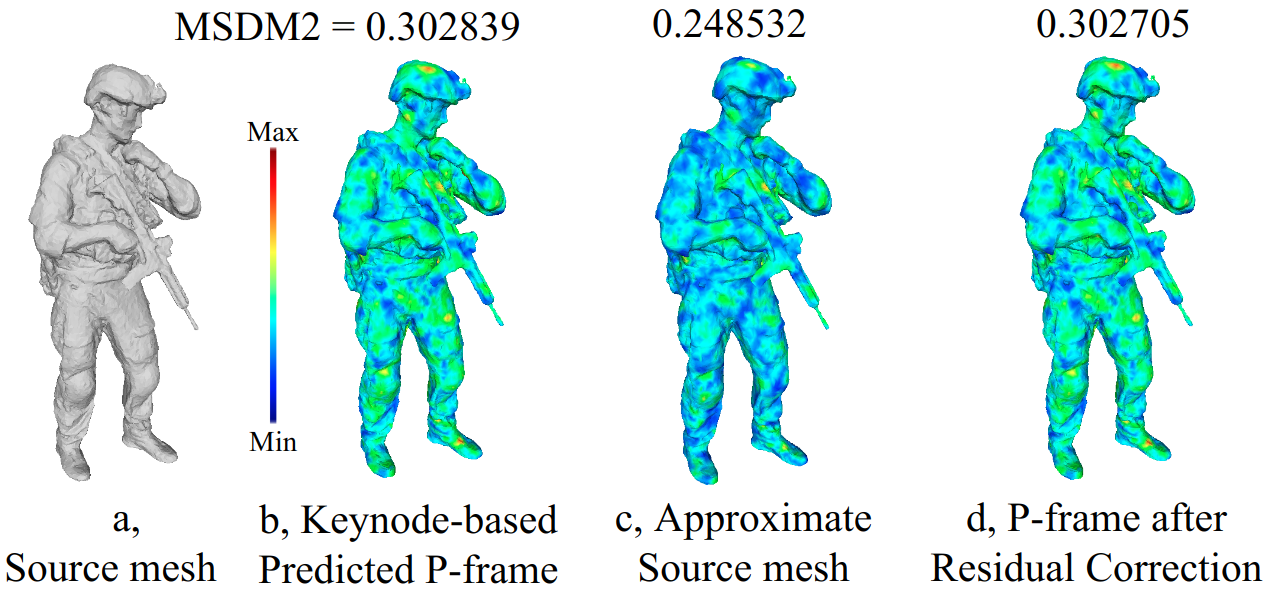}
\caption{MSDM2 Distortion at each stage of compressing frame 537 in \textit{Soldier} as the first P-frame in a GoF.}
\label{fig:errors-each-step}
\end{figure}

Fig. \ref{fig:errors-each-step} shows the geometric distortion at each compression phase of our method, specifically: the predicted P-frame prior to residual coding (b), the approximate source mesh for that frame (c), and the corrected mesh post-residual coding (d).

In Fig. \ref{fig:errors-each-step}b, the frame predicted based on key nodes shows some degree of distortion, which is then reduced with residual coding, as seen in Fig. \ref{fig:errors-each-step}d. The approximate source mesh in Fig. \ref{fig:errors-each-step}c serves as a reference for the residual calculation. In essence, if the residuals could be compressed losslessly, Fig. \ref{fig:errors-each-step}c would represent the best-corrected frame.

\subsection{Influence of GoF Size and I-frame Quality}

In this section, we evaluate various GoF sizes and I-frame qualities to understand how they impact compression performance. The experiment is done on \textit{Thomas}'s first 32 frames, with five different GoF sizes and four different I-frame quality levels. We evaluate the I-frame quality levels 1 to 4 from lower to higher. For P-frames, we use the same encoding configuration across all cases being assessed,  and the prediction mode is Fully Forward.

\begin{figure}[htb]
\centering
\includegraphics[width=0.465\textwidth]{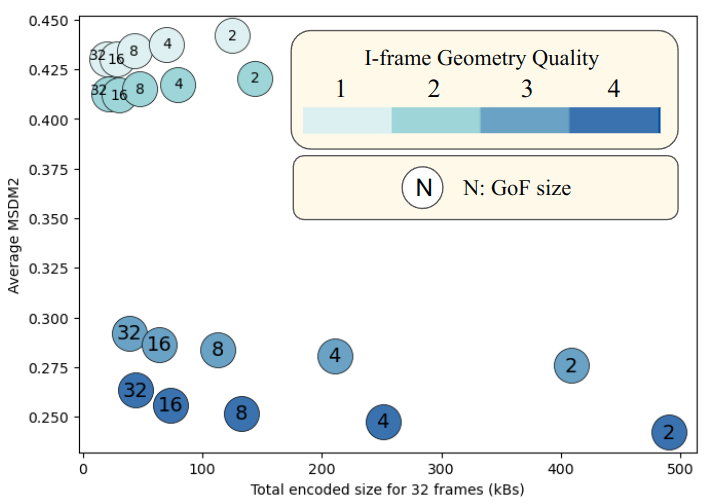}
\caption{Influence of GoF size and I-frame quality on compression performance - first 32 frames in \textit{Thomas} sequence.}
\label{fig:influence-GoF-size-IF-quality}
\end{figure}

Fig.~\ref{fig:influence-GoF-size-IF-quality} shows that smaller GoF sizes are more sensitive to variations in I-frame quality, as indicated by a wider spread in compression performance when changing quality levels. Smaller GoF sizes lead to higher bit rates, whereas larger GoF sizes tend to perform better at lower bit rates. For instance, at similar quality settings, GoF size 16 achieves a lower bit rate. Overall, Fig.~\ref{fig:influence-GoF-size-IF-quality} suggests that the KeyNode-based method prefers medium to high I-frame quality. When aiming for lower target bit rates, it is advantageous to increase the GoF size rather than reduce the quality of the I-frames.

Additionally, it can be noticed that when the I-frame geometry quality is low (e.g., 1 and 2), increasing the GoF size slightly reduces distortion. This occurs because the octree-based residual coding effectively enhances the quality of P-frames, causing P-frames to have slightly lower distortion than I-frames.

\subsection{Influence of P-frame Encoding Components}

\begin{figure}[htb]
\centering
\includegraphics[width=0.485\textwidth]{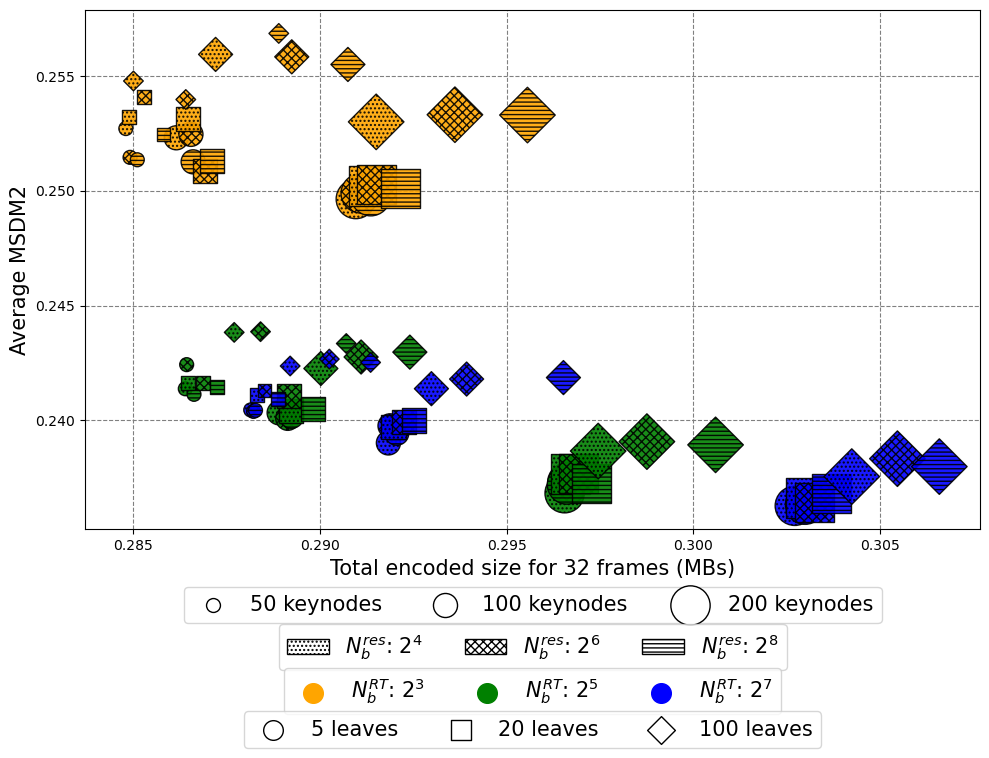}
\caption{Influence of P-frame compression parameters for the first ten frames of {Mitch} with GoF size = 4.} 
\label{fig:param-influence-Thomas}
\end{figure}

P-frames are predicted and encoded using the key nodes and their transformations RT, as well as the geometric residuals. The values chosen for each component affect the compression performance.
To assess sensitivity to variations in parameters, we examine performance across many parameter sets in Fig.~\ref{fig:param-influence-Thomas}. 

The dominant factor influencing overall quality is the precision of motion transformations, determined by the number of quantization bins $N^{RT}_{b}$. When $N^{RT}_{b}$ is low (e.g., $2^3$, orange markers), motion prediction becomes inaccurate, leading to high distortion. In this case, increasing the number of key nodes, refining octree size, or improving residual quantization cannot compensate for the poor motion modeling, highlighting the importance of precise motion representation.

Once $N^{RT}_{b}$ is sufficiently large (e.g., $2^5$ or $2^7$, green and blue markers), increasing the number of key nodes slightly improves distortion by refining motion granularity. Finer octree subdivisions can enhance adaptation to local residual structures, as observed in cases such as the 50-keynode green markers and 100-keynode orange markers with $N^{res}_{b}=2^6$ (cross-hatch pattern), where increasing the octree size from 5 to 20 leaf nodes reduces distortion. However, excessive partitioning—particularly in minimal residual scenarios such as the \textit{Mitch} sequence—can sometimes disrupt residual coherence and amplify minor artifacts and unsmooth details, slightly degrading perceptual quality even though vertex positions improve in a least-squares sense. Finally, the influence of residual quantization $N^{res}_{b}$ is subtle when octree sizes are small but becomes increasingly pronounced as finer spatial subdivisions amplify the bitrate sensitivity to residual precision.

The performance difference between $N^{RT}_{b} = 2^5$ and $2^7$ is small, suggesting that once a sufficient precision is reached for motion transformations, additional refinement offers diminishing returns.

\subsection{Compression Performance Comparison}

In this section, we compare the Rate-Distortion (RD) performance of our KeyNode-based method (under \textit{ADP} mode) with the state-of-the-art V-DMC's geometry coding method \cite{9922888} with their two modes: all intra-coding (denoted by \textit{V-DMC - intra}), and with inter coding frames (denoted by \textit{V-DMC - inter}). V-DMC was chosen as the baseline because it demonstrated the highest effectiveness among methods responding to the MPEG call for proposals and offers a publicly available implementation, enabling fair and reproducible comparisons.

\begin{figure}[htb]
\centering
\includegraphics[width=0.45\textwidth]{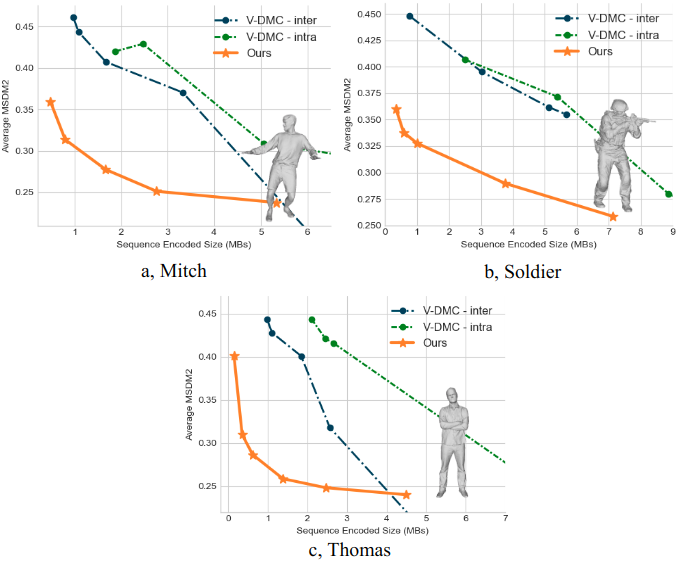}
\caption{Geometric distortion comparison of methods on Slow Motion sequences.} 
\label{fig:RD-MSDM2-cat1}
\end{figure}

For Slow Motion sequences, Fig.~\ref{fig:RD-MSDM2-cat1} presents the RD curves using the MSDM2 metric.
 The results show that our method outperforms V-DMC considerably on these sequences, especially at low bitrates.

\begin{figure}[htb]
\centering
\includegraphics[width=0.44\textwidth]{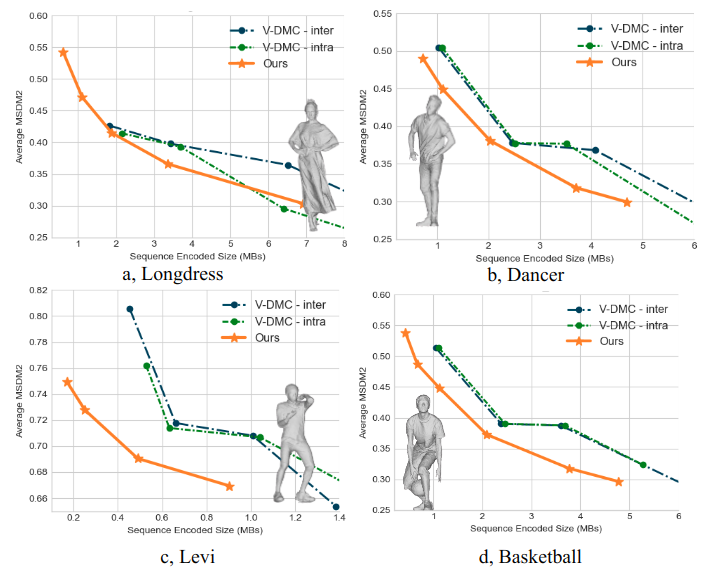}
\caption{Geometric distortion comparison on Fast Motion and Join/Split categories.}
\label{fig:RD-MSDM2-cat23}
\end{figure}

For the Fast Motion and Join/Split categories, Fig.~\ref{fig:RD-MSDM2-cat23} depicts the RD curves using MSDM2. Our method is slightly better than V-DMC on \textit{Longdress}, and notably outperforms V-DMC on \textit{Dancer}, \textit{Levi}, and \textit{Basketball}.  

The \textit{Basketball} sequence involves different objects joining and separating over time. Since our method uses a fixed size for the GoF throughout the sequence, our prediction model may encounter challenges when objects join and separate. The compression performance might be improved by incorporating an adaptive GoF scheme, allowing for the dynamic insertion of I-frames as needed.
 
The curves in Figures \ref{fig:RD-MSDM2-cat1} and \ref{fig:RD-MSDM2-cat23} can be summarized in terms of average bitrate savings using Bjøntegaard Delta rate (BD-rate) percentage, following the methodology in  \cite{Bjntegaard2001CalculationOA}. 
BD-rate measures the difference in bit rate required to achieve the same level of distortion between two compression methods. A lower BD-rate indicates better compression efficiency, as it achieves the same quality with less data. 
Table \ref{tab:BD-rate-cat1} presents the BD-rate percentages compared to two coding modes of V-DMC; negative values indicate that Ours outperforms the comparison method, while positive values indicate that Ours performs worse.
According to Table \ref{tab:BD-rate-cat1}, our methods considerably outperform V-DMC for sequences in all categories.

\begin{table}[]
\caption{BD-rate (\%) of Ours against V-DMC.}
\centering
\renewcommand{\arraystretch}{1.5} 
\resizebox{0.498\textwidth}{!}{
\begin{tabular}{c|ccc|ccc|c|l}
\hline
\multirow{2}{*}{} & \multicolumn{3}{c|}{\textbf{Slow Motion}}                                                    & \multicolumn{3}{c|}{\textbf{Fast Motion}}                                                     & \textbf{Join/Split} & \multicolumn{1}{c}{\multirow{2}{*}{\textbf{Avg.}}} \\ \cline{2-8}
                  & \multicolumn{1}{c|}{Soldier}        & \multicolumn{1}{c|}{Thomas}          & Mitch           & \multicolumn{1}{c|}{Longdress}       & \multicolumn{1}{c|}{Dancer}          & Levi            & Basketball          & \multicolumn{1}{c}{}                               \\ \hline
V-DMC inter       & \multicolumn{1}{c|}{\textbf{-93.6}} & \multicolumn{1}{c|}{\textbf{-91.55}} & \textbf{-87.55} & \multicolumn{1}{c|}{\textbf{-36.78}} & \multicolumn{1}{c|}{\textbf{-24.55}} & \textbf{-58.57} & \textbf{-34.37}     & -60.99                                             \\ \hline
V-DMC intra       & \multicolumn{1}{c|}{\textbf{-77.3}} & \multicolumn{1}{c|}{\textbf{-94.66}} & \textbf{-82.97} & \multicolumn{1}{c|}{\textbf{-18.74}} & \multicolumn{1}{c|}{\textbf{-23.62}} & \textbf{-59.14} & \textbf{-34.71}     & -55.87                                             \\ \hline
Average           & \multicolumn{3}{c|}{-87.94}                                                                  & \multicolumn{3}{c|}{-36.9}                                                                    & -34.54              & \textbf{-58.43}                                    \\ \hline
\end{tabular}}
\label{tab:BD-rate-cat1}
\end{table}

\subsection{Ablation studies}
\begin{figure}[htb]
\centering
\includegraphics[width=0.49\textwidth]{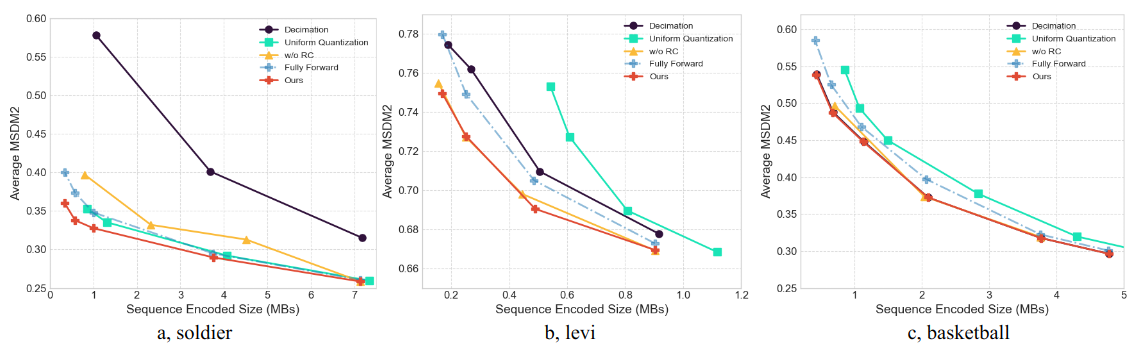}
\caption{Ablation study for three representative sequences for three categories.}
\label{fig:ablation}
\end{figure}

We conducted a leave-one-out ablation study for three representative sequences (\textit{Soldier} for slow-motion, \textit{Levi} for fast-motion, and \textit{Basketball} for join-split category), where we begin with the full method and disable one component at a time. This allows us to assess the individual contribution of each module by observing the performance drop when it is removed. The ablation, shown in Figure \ref{fig:ablation}, includes:

\begin{itemize}[]
    \item \textit{Decimation:} Replace our key node generator with mesh decimation.
    \item \textit{Uniform Quantization:} Replace our Cauchy-based quantization and coding with uniform quantization and explicit Huffman tables for encoding of Transformations RT.
    \item \textit{Fully Forward:} Replace adaptive dual-direction prediction with a fixed forward prediction.
    \item \textit{w/o RC:} Disable residual coding.
    \item \textit{Ours:} Full method with all components enabled.
\end{itemize}

Figures \ref{fig:ablation}a and \ref{fig:ablation}b show that for the \textit{Levi} and \textit{Soldier} sequences, our Optimal Key Node Generator achieves significantly lower distortion compared to Mesh Decimation. However, for \textit{Basketball} in Fig. \ref{fig:ablation}c, the improvement is less pronounced, with the curves overlapping. This is likely due to the use of a search step size (20-200 nodes per step) in our generator for computational efficiency, which yields near-optimal rather than globally optimal key nodes. A finer search step would likely enhance performance further.
Additionally, the large number of key nodes selected for \textit{Basketball} (300 vs. 100 for \textit{Levi} and \textit{Soldier}) may diminish the visible difference between methods, suggesting that \textit{Mesh Decimation} may suffice in scenarios with a large key node count, thus reducing computational complexity.
Overall, the \textit{Decimation} curves emphasize the effectiveness of our KeyNode-driven codec, particularly in capturing temporal motion in slow-motion sequences, while our Optimal Key Node Generator excels at selecting the right set of key nodes for this purpose.

For all categories, the \textit{Uniform Quantization} curves show that the bitrate needed is significantly higher without fitting into a Cauchy distribution to convey the Huffman tables implicitly. 

The \textit{w/o RC} curves indicate that residual coding is particularly effective in the slow-motion category. This may seem counterintuitive when compared to traditional 2D video coding, where residual coding typically yields greater improvements in fast-motion sequences.  In 2D video, pixel-to-pixel correspondence is explicit, but in scanned meshes, this correspondence is not available, and our residual estimation method is less effective for complex-motion sequences. As a result, even with perfect residual correction, the quality improvement is either small or negative, and therefore, Residual Coding is disabled for multiple frames. This suggests that improving residual estimation in complex-motion scenarios could be a promising area for future work.

 \subsection{Frame-wise and GoF-wise Distortion Analysis}

\begin{figure}[h]
\centering
\includegraphics[width=0.49\textwidth]{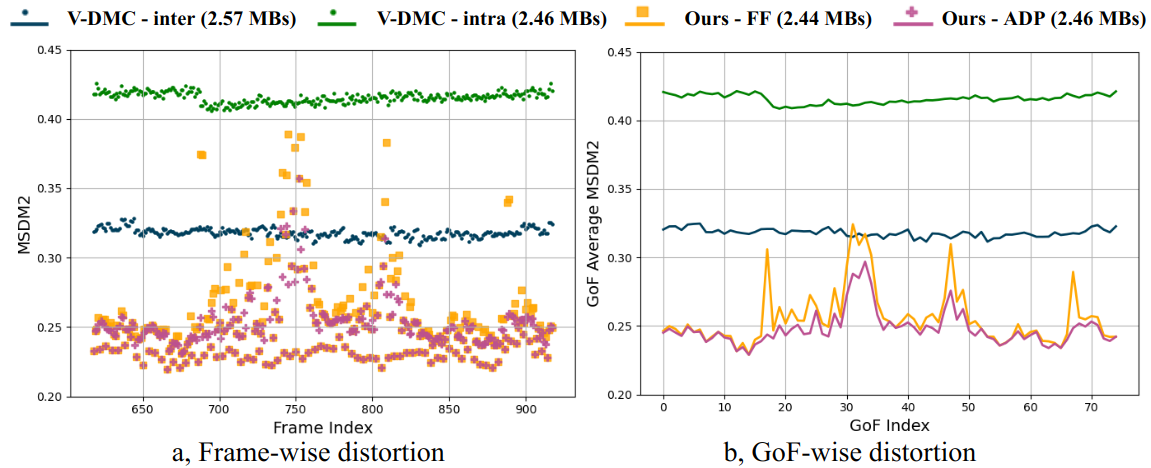}
\caption{Frame-wise (a) and GoF-wise (b) analysis for \textit{Thomas} under similar sequence encoded size. The evaluated GoF size is 4 frames. \textit{Label format: Method (Sequence encoded size)}.}
\label{fig:frame-wise-GoF-wise-PSNR-Thomas}
\end{figure}

Fig.~\ref{fig:frame-wise-GoF-wise-PSNR-Thomas} provides frame-wise and GoF-wise distortion for different methods at a similar encoded size. In Fig.~\ref{fig:frame-wise-GoF-wise-PSNR-Thomas}a, among our KeyNode-based coded frames in both FF and ADP modes, the I-frames exhibit the smallest errors, as they are encoded with high quality using a static mesh encoder. As seen in Fig.~\ref{fig:frame-wise-GoF-wise-PSNR-Thomas}b, our FF mode typically exceeds V-DMC's performance across most GoFs, although there is one GoF in the middle of the sequence where it performs worse than V-DMC's inter-coding. However, our ADP mode manages this GoF effectively.

\begin{figure}[htb]
\centering
\includegraphics[width=0.5\textwidth]{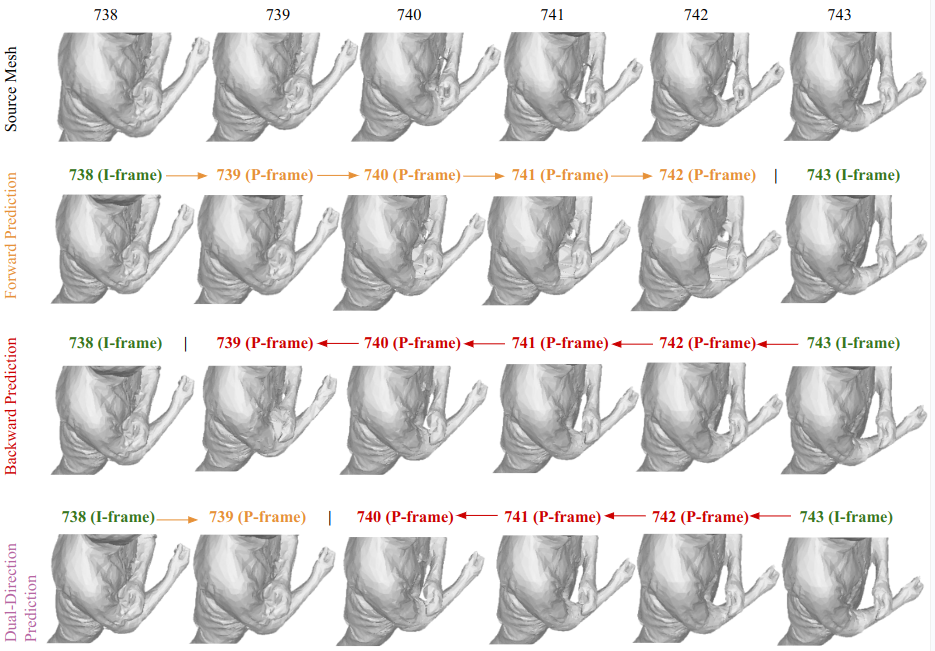}
\caption{Decoded meshes illustrate the topological change issue in \textit{Thomas}, managed by Dual-direction prediction.}
\label{fig:Thomas-topo-change}
\end{figure}

To examine that specific GoF more closely, Fig.~\ref{fig:Thomas-topo-change} displays the frames within this GoF encoded using different modes. In the source mesh, frames 738 and 739 show the left arm attached to the body, while frames 740 to 743 depict the arm gradually detaching from the body, indicating a topological change.
It is clear that forward prediction struggles to capture the arm's detachment accurately. In contrast, the third row, which uses backward prediction, shows that frames 740, 741, and 742 are accurately predicted from I-frame 743, as they preserve a similar topology.
With Dual-direction Prediction, which leverages both prediction directions, P-frames within the GoF can be predicted more accurately using I-frames with a similar, though not necessarily identical, topological structure. The visualizations effectively demonstrate how ADP mode reduces distortion in this GoF.

\begin{figure}[h]
\centering
\includegraphics[width=0.45\textwidth]{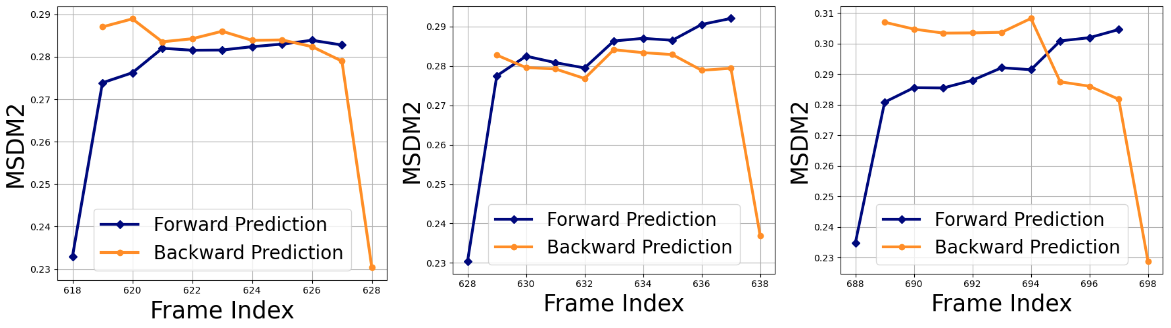}
\caption{Frame-wise MSDM2 from two prediction directions in three GoFs, where the ideal direction change index is not in the middle of the GoF, highlighting the preference for an adaptive scheme.}
\label{fig:dual-direction-PSNR}
\end{figure}

Besides the example in Fig.~\ref{fig:Thomas-topo-change}, where it is advantageous to switch to backward prediction at a frame that is not the middle of the GoF, Fig.~\ref{fig:dual-direction-PSNR} shows additional examples to demonstrate the benefits of an adaptive scheme for direction switching, since backward prediction could perform better not only in the middle of the GoF but also near its beginning or end.

\subsection{Geometry Reconstruction Distortion}

\begin{figure}[htb]
\centering
\includegraphics[width=0.48\textwidth]{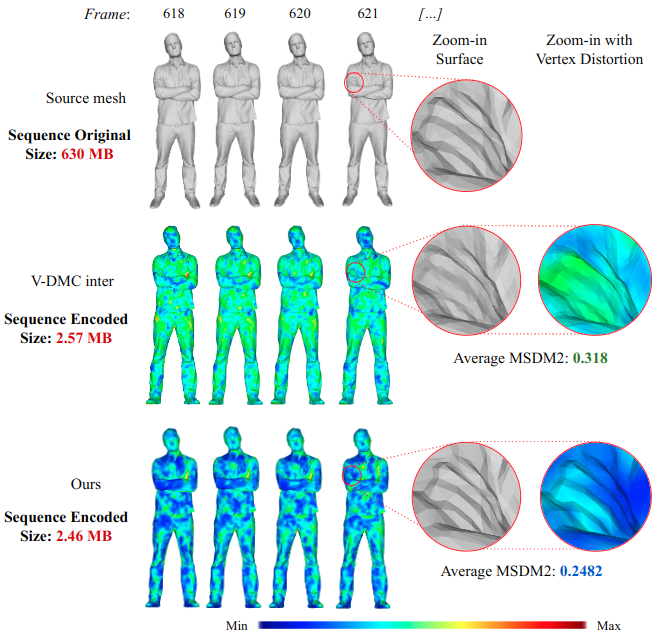}
\caption{Reconstruction distortion of our method and V-DMC, under similar bitrate for \textit{Thomas}.}
\label{fig:Thomas-comparable-frames}
\end{figure}

Fig.~\ref{fig:Thomas-comparable-frames} provides a visual comparison between the reconstruction errors of our compression approach and V-DMC under similar bitrate for the \textit{Thomas} sequence. The meshes are colored based on MSDM2 distortion, indicating the accuracy of geometry reconstruction. 
Warmer colors (reds) indicate higher reconstruction errors, while cooler colors suggest areas where the reconstruction closely matches the original mesh. 

Notably, with \textit{Thomas} in Fig.~\ref{fig:Thomas-comparable-frames}, our method achieves lower errors with a similar bitrate. In the zoom-in region, our decoded mesh more closely resembles the original mesh than that of V-DMC, which tends to lose the details of the fingers, despite appearing smoother due to the increased number of vertices from heavy subdivision.

\begin{figure}[htb]
\centering
\includegraphics[width=0.485\textwidth]{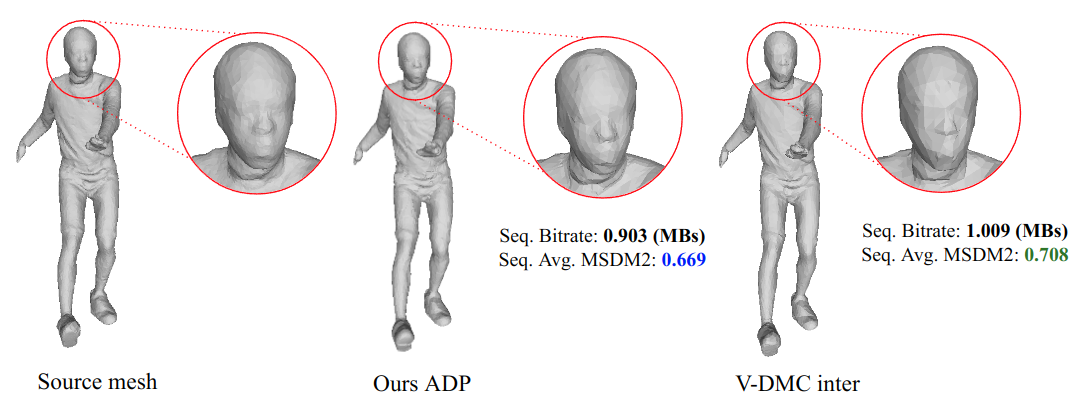}
\caption{Frame 145 of \textit{levi}, extracted from the decoded sequences using our method and V-DMC.}
\label{fig:levi-comparable-frames}
\end{figure}

Similarly, Fig.~\ref{fig:levi-comparable-frames} provides a visual comparison of a frame extracted from decoded \textit{Levi}  sequences. The total sequence encoded bitrate and average MSDM2 are listed alongside each method.
Fig.~\ref{fig:levi-comparable-frames} shows that for that decoded frame, our method preserves details more effectively than V-DMC. This is particularly noticeable when zooming in on the model's head, where our decoded mesh maintains the shape of the nose and lips more accurately.

\subsection{Theoretical Complexity Analysis}
In this section, we evaluate the theoretical complexity of our method and V-DMC. Since the coding of I-frames is identical in both methods, we focus on comparing the complexity of the inter-mode, which is used for encoding predictive frames.
It is important to note that the complexity estimates are based on theoretical concepts outlined in related papers, and the real-world complexity may vary depending on the actual implementation.

Tables \ref{tab:VDMC-complexity} and \ref{tab:our-complexity} summarize the estimated computational complexity of V-DMC and our method, respectively, where $V$ is the number of vertices in the mesh.
The analysis shows that both methods exhibit comparable theoretical complexity when the number of subdivided vertices in V-DMC satisfies $S \leq V$. However, in scenarios where $S > V$, its computational cost can grow significantly. In contrast, our method maintains a more stable complexity profile by directly operating at the vertex level.

\begin{table}[h]
\caption{V-DMC's theoretical complexity. $B$: number of vertices in the base mesh; $S$: number of vertices in the subdivided mesh.}
\centering
\resizebox{0.4\textwidth}{!}{
\begin{tabular}{lc|cc}
\hline
\multicolumn{2}{l|}{\textbf{Component}}                          & \multicolumn{1}{c|}{\textbf{Best-case}}     & \textbf{Worst-case} \\ \hline
\multicolumn{2}{l|}{Pre-processing}           & \multicolumn{1}{c|}{$O(V \log V)$} & $O(V^2)$          \\ \hline
\multicolumn{2}{l|}{Base Mesh Reuse Decision} & \multicolumn{2}{c}{$O(S \log S)$}                      \\ \hline
\multicolumn{2}{l|}{Motion Field Encoder}     & \multicolumn{2}{c}{$O(B)$}                             \\ \hline
\multicolumn{2}{l|}{Displacement Coding}      & \multicolumn{1}{c|}{$O(S \log S)$} & $O(S^2)$          \\ \hline
\multicolumn{1}{l|}{\multirow{2}{*}{\textbf{Total}}} & if $V \geq S$ & \multicolumn{1}{c|}{\textbf{$O(V \log V)$}} & \textbf{$O(V^2)$}   \\ \cline{2-4} 
\multicolumn{1}{l|}{}       & if $V < S$      & \multicolumn{1}{c|}{\textbf{$O(S \log S)$}} & \textbf{$O(S^2)$}   \\ \hline
\end{tabular}}
\label{tab:VDMC-complexity}
\end{table}

\begin{table}[h]
\caption{Our theoretical complexity. $N$: number of key nodes.}
\centering
\resizebox{0.4\textwidth}{!}{
\begin{tabular}{ll|cc}
\hline
\multicolumn{2}{l|}{\textbf{Component}} & \multicolumn{1}{c|}{\textbf{Best-case}} & \textbf{Worst-case} \\ \hline
\multicolumn{2}{l|}{Key Node Generation}      & \multicolumn{2}{c}{$O(V \log V)$}             \\ \hline
\multicolumn{2}{l|}{Transformation Extractor} & \multicolumn{2}{c}{$O(V \log V)$}             \\ \hline
\multicolumn{2}{l|}{Key Node Encoder}         & \multicolumn{2}{c}{$O(N \log V)$}             \\ \hline
\multicolumn{2}{l|}{Transformation Encoder}   & \multicolumn{2}{c}{$O(N)$}                    \\ \hline
\multicolumn{2}{l|}{Residual Coding}          & \multicolumn{1}{c|}{$O(V)$}        & $O(V^2)$ \\ \hline
\multicolumn{2}{l|}{\textbf{Total}}           & \multicolumn{1}{c|}{$O(V \log V)$} & $O(V^2)$ \\ \hline
\end{tabular}}
\label{tab:our-complexity}
\end{table}

\section{Discussion and Future Work}

In this paper, we introduce an efficient approach to compress real-world scanned 3D dynamic human meshes by leveraging embedded key nodes and their transformations to capture the temporal evolution of vertices. We proposed a strategy where the temporal change of each vertex is computed as a weighted combination of key node transformations, facilitating a comprehensive representation of motion. Our KeyNode-driven method addresses the challenges of dynamic human mesh compression by effectively handling topology changes between frames and accommodating scan imperfections. Through the use of sparse key nodes, our approach accurately captures and represents the complexities of human motion, achieving efficient compression. To enhance the quality of the KeyNode-based prediction, we present an octree-based residual coding scheme and a Dual-direction prediction mode to further improve temporal prediction.

Our method presents notable advantages in geometry coding compared to V-DMC, particularly evident in the Slow Motion sequence category. For Fast Motion and Join/Split sequences,  the improvements are smaller but still considerable. These results highlight the effectiveness of our method in modeling the motion while handling varying topology and scan defects. It also emphasizes the challenge of accurately predicting the geometry of P-frames in human dynamic meshes with rapid motions.

Our KeyNode-driven codec introduces a new way to compress scanned 3D human dynamic models, opening up numerous avenues for future research and development. One promising direction is the adaptive placement of I-frames, where dynamically determining the optimal points for I-frame insertion could further enhance compression performance. Another potential improvement is the integration of view-dependent coding, which takes into account that users typically view a 3D model from a single angle at a time. By reducing the quality of unseen areas, we can lower bitrate requirements without compromising the user experience. Additionally, while the current KeyNode-driven codec delivers effective results, its encoding complexity remains a limitation. The optimization-based nature of components such as the \textit{Optimal Key Node Generator} and \textit{Rotation and Translation Extractor} can be computationally demanding. Future work could focus on relaxing these optimization steps and exploring parallelization techniques to improve efficiency. Furthermore, the KeyNode-driven codec holds great potential for broader applicability by adapting the prediction model to align with the characteristics of diverse mesh structures. While the current method is optimized for objects exhibiting a combination of rigid and non-rigid motions, such as dynamic human meshes, future work could explore extending the transformation types, e.g., adding Scaling and Shearing in addition to Rotation and Translation, unlocking its versatility across a range of scanned models.

\bibliographystyle{IEEEtran}
\bibliography{egbib}

\vfill

\end{document}